\documentclass[sigconf]{acmart}
\usepackage{booktabs}
\usepackage{multirow}
\usepackage{array}
\usepackage{hyperref}
\usepackage{url}
\usepackage{nicefrac}       
\usepackage{microtype}      
\usepackage{xcolor}         

\usepackage{wrapfig}
\usepackage{threeparttable}
\usepackage{comment}
\usepackage{multicol}

\usepackage{amsmath,amsthm,amsfonts,amssymb}
\usepackage{mathtools}
\usepackage{algorithm}
\usepackage{algorithmic}
\usepackage{tkz-euclide} 
\usepackage{siunitx}
\usepackage{bm}
\usepackage{subcaption,graphicx}
\usepackage{pifont}
\usepackage{arydshln}
\usepackage{makecell}
\usepackage{textcomp}
\theoremstyle{plain}

\theoremstyle{definition}

\theoremstyle{remark}

\newcommand{\nn}{\num[group-separator={,},group-minimum-digits=3]}
\newcommand{\B}[1]{\textbf{#1}}

\AtBeginDocument{%
	}

\copyrightyear{2025}
\acmYear{2025}
\setcopyright{acmlicensed}
\acmConference[MM '25]{Proceedings of the 33rd ACM International Conference on Multimedia}{October 27--31, 2025}{Dublin, Ireland}
\acmBooktitle{Proceedings of the 33rd ACM International Conference on Multimedia (MM '25), October 27--31, 2025, Dublin, Ireland}
\acmDOI{10.1145/3746027.3755207}
\acmISBN{979-8-4007-2035-2/2025/10}
\acmSubmissionID{2606}

\begin{document}
	
	\title{RealHD: A High-Quality Dataset for Robust Detection of State-of-the-Art AI-Generated Images}
	
	\author{Hanzhe Yu}
	\email{yuhanzheyyy@163.com}
	\orcid{0009-0008-5791-0546}
	\affiliation{
		\department{Institute of Cyberspace Security,}
		\institution{Zhejiang University of Technology}
		\city{Hangzhou}
		\country{China}
	}
	
	\author{Yun Ye}
	\email{yun.ye@intel.com}
	\orcid{0000-0002-1905-4354}
	\affiliation{
		\institution{Intel Corporation}
		\city{Beijing}
		\country{China}
	}

	\author{Jintao Rong}
	\email{2111903071@zjut.edu.cn}
	\orcid{0000-0002-3173-757X}
	\affiliation{
		\department{College of Information Engineering,}
		\institution{Zhejiang University of Technology}
		\city{Hangzhou}
		\country{China}
	}

	\author{Qi Xuan}
	\email{xuanqi@zjut.edu.cn}
	\orcid{0000-0002-1042-470X}
	\affiliation{
		\department{Institute of Cyberspace Security,}
		\institution{Zhejiang University of Technology}
		\city{Hangzhou}
		\country{China}
	}
	\affiliation{
		\institution{Binjiang Institute of Artificial Intelligence, ZJUT}
		\city{Hangzhou}
		\country{China}
	}
	
	\author{Chen Ma}
	\email{machen@zjut.edu.cn}
	\orcid{0000-0001-6876-3117}
	\authornote{Corresponding author.}
	\affiliation{
	\department{Institute of Cyberspace Security,}
	\institution{Zhejiang University of Technology}
	\city{Hangzhou}
	\country{China}
	}
	\affiliation{
		\institution{Binjiang Institute of Artificial Intelligence, ZJUT}
		\city{Hangzhou}
		\country{China}
	}
	
	\renewcommand{\shortauthors}{Hanzhe Yu, Yun Ye, Jintao Rong, Qi Xuan, \& Chen Ma}
	\begin{abstract}
		The rapid advancement of generative AI has raised concerns about the authenticity of digital images, as highly realistic fake images can now be generated at low cost, potentially increasing societal risks. In response, several datasets have been established to train detection models aimed at distinguishing AI-generated images from real ones. However, existing datasets suffer from limited generalization, low image quality, overly simple prompts, and insufficient image diversity.
		To address these limitations, we propose a high-quality, large-scale dataset comprising over \nn{730000} images across multiple categories, including both real and AI-generated images. The generated images are synthesized via state-of-the-art methods, including text-to-image generation (guided by over \nn{10000} carefully designed prompts), image inpainting, image refinement, and face swapping. Each generated image is annotated with its generation method and category. Inpainting images further include binary masks to indicate inpainted regions, providing rich metadata for analysis. Compared to existing datasets, detection models trained on our dataset demonstrate superior generalization capabilities. Our dataset not only serves as a strong benchmark for evaluating detection methods but also contributes to advancing the robustness of AI-generated image detection techniques. Building upon this, we propose a lightweight detection method based on image noise entropy, which transforms the original image into an entropy tensor of Non-Local Means (NLM) noise before classification. Extensive experiments demonstrate that models trained on our dataset achieve strong generalization, and our method delivers competitive performance, establishing a solid baseline for future research. The dataset and source code are publicly available at \url{https://real-hd.github.io}.
	\end{abstract}

	
	
	\begin{teaserfigure}
		\centering
		\includegraphics[width=\textwidth]{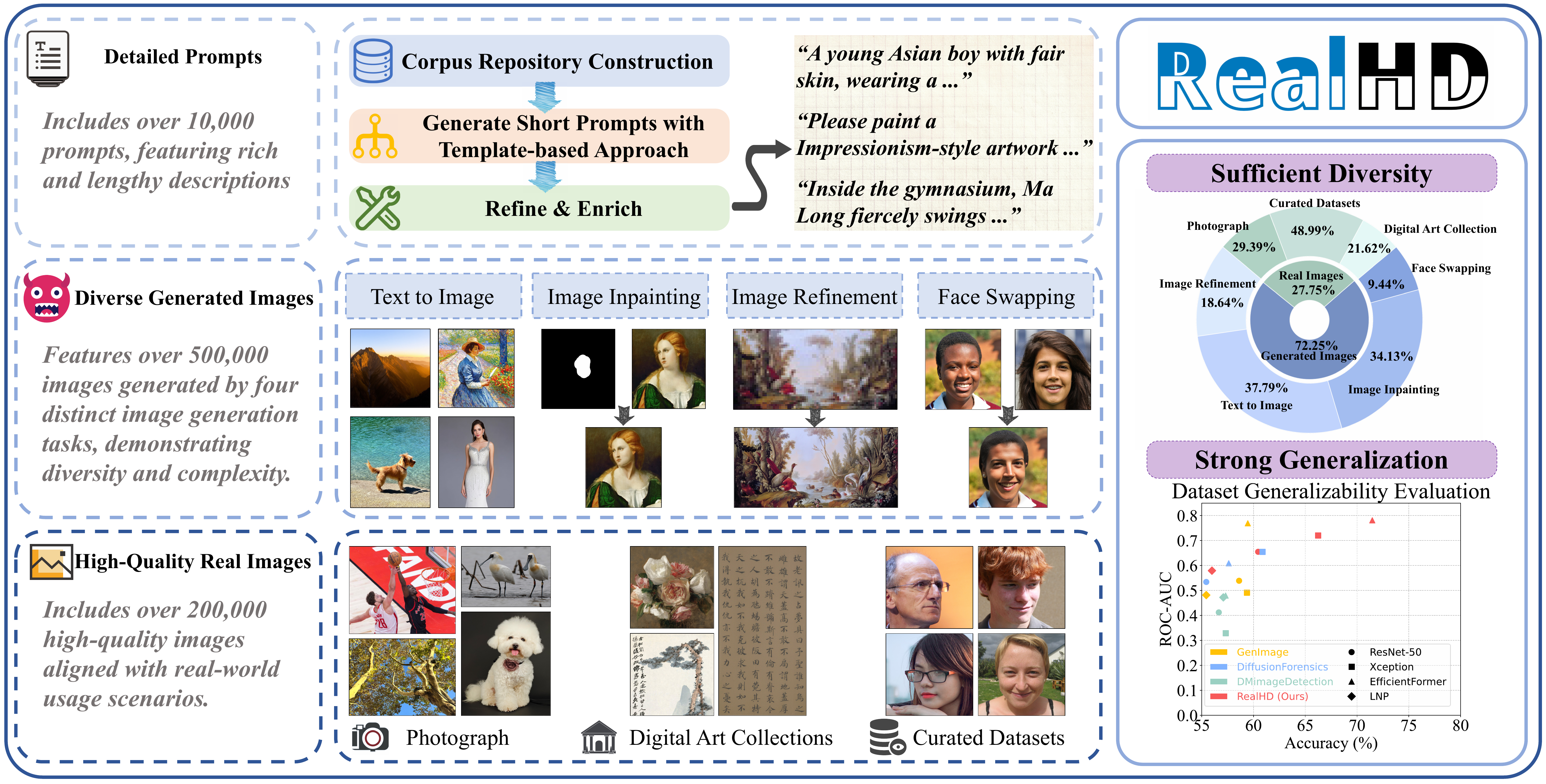}
        \vspace{-3mm}
        \Description{Overview of the RealHD dataset showing diverse prompts, generated images, and high-quality real images across multiple themes.}
		\caption{Overview of the RealHD Dataset. The RealHD dataset features a diverse set of detailed prompts, images generated by various methods, and carefully curated high-quality real images covering multiple themes. By combining high quality with broad diversity, training on RealHD significantly enhances model generalization for real-world detection tasks.}
		\label{fig:teaser}
	\end{teaserfigure}

	\maketitle

\section{Introduction}
	The rapid advancement of generative models has led to the emergence of various high-quality, open-source models, such as Stable Diffusion \cite{arombach2022high} and Flux \cite{flux2024,labs2025flux1kontextflowmatching}, which have made generated images increasingly realistic and natural. These models democratize creative expression by equipping people without formal artistic training with advanced generative tools. However, the widespread availability of such models raises concerns about image authenticity, as malicious actors can easily create forged images, posing challenges to image forensics \cite{ferreira2020review}, misinformation governance \cite{xu2023combating}, and copyright protection \cite{ren2024copyright}, with serious social consequences. In response, the community has attempted to construct high-quality image datasets to distinguish fake images from real ones. However, current datasets have limitations in real-world scenarios. For instance, UADFV \cite{li2018UADFV}, ForgeryNet \cite{he2021forgerynet}, and DiFF \cite{cheng2024diffusion} focus on facial forgery detection, making them difficult to generalize to other common scenarios. DE-FAKE \cite{sha2022defake}, CIFAKE \cite{bird2024cifake}, DiffusionForensics~\cite{wang2023dire}, and Chameleon \cite{yan2025sanity} are datasets that incorporate diffusion-generated images to support detection tasks. 
	While some of these datasets provide relatively high-resolution images, they are typically limited in size (thousands to tens of thousands of images), which constrains the generalization of detection models to broader real-world scenarios. In contrast, GenImage \cite{zhu2023genimage} builds a large-scale dataset containing millions of AI-generated images. 
	However, the real images in GenImage, obtained from ImageNet \cite{deng2009imagenet}, suffer from low visual quality due to compression artifacts, and their categories do not align with the application scenarios of generative models.
	 Additionally, existing datasets primarily focus on text-to-image or image-to-image generation, while lacking coverage of more diverse generative tasks such as image inpainting or image refinement.
	
	To overcome these challenges and enhance the generalization of detection, we introduce RealHD, a high-quality dataset that is well-suited to real-world application scenarios (Fig. \ref{fig:teaser}). The real images in RealHD consist of lossless and high-quality compressed JPEG images with quality $\geq 90$, collected from multiple sources and diverse themes. 
	For generated images, we select image generation types that are either prevalent on social media or associated with significant ethical and safety risks, and craft detailed prompts based on the categories of real images to produce high-quality outputs. All synthetic images are generated using state-of-the-art diffusion models (Stable Diffusion v1.5 to v3.5, FLUX, and others) and are reviewed by human annotators.
	Notably, experimental results demonstrate that models trained on our dataset outperform those trained on existing datasets when evaluated on the Chameleon dataset \cite{yan2025sanity} (where all images pass the human-perception ``Turing test''), highlighting its superior generalization capability.
	
	Building upon RealHD, we further establish a simple yet effective baseline for AI-generated image detection. To assess the dataset and expose its challenges, we evaluate four existing detectors \cite{he2016deep,chollet2017xception,li2022efficientformer,liu2022detecting} and uncover significant generalization bottlenecks across diverse generative modalities. For instance, Xception~\cite{chollet2017xception} achieves 99.00\% accuracy and an Area Under the Receiver Operating Characteristic Curve (AUC) of 0.9996 on the GenImage dataset~\cite{zhu2023genimage}, which contains only text-to-image samples, but drops to 85.21\% accuracy and 0.9515 AUC on the RealHD dataset, which incorporates multiple generative modalities. To address this limitation, we propose a lightweight detection method leveraging the entropy of low-level noise patterns---features that end-to-end RGB networks struggle to extract. Our method achieves 15.90\% relative improvement in accuracy and 5.00\% relative improvement in AUC on the Xception network, validating its effectiveness as a baseline.
	 \begin{table*}
		\centering
		\small
		\caption{Overview of image generation datasets. T2I denotes text-to-image generation, INP denotes image inpainting, REF denotes image refinement, and FS denotes face swapping.}
		\label{tab:datasets_overview}
		\begin{tabular}{lcccccccl}
			\toprule
			\multirow{2}{*}{\B{Dataset}}
			&\multirow{2}{*}{\B{Public Availability}} 
			& \multicolumn{4}{c}{\B{Methodology}} 
			& \multirow{2}{*}{\B{Real Images}}
			& \multirow{2}{*}{\B{Generated Images}}
			& \multirow{2}{*}{\B{Dataset Type}} \\
			\cmidrule(lr){3-6} 
			& & T2I & INP & REF & FS & & & \\
			\midrule
			UADFV \cite{li2018UADFV} & \ding{51} &\ding{55} & \ding{55} & \ding{55} &\ding{51} & \nn{17329} & \nn{16991} &Face forgery\\
			ForgeryNet \cite{he2021forgerynet} & \ding{51} & \ding{55}& \ding{55}& \ding{55}& \ding{51} &\nn{1438201} & \nn{1457861}&Face forgery\\
			DiFF \cite{cheng2024diffusion} & \ding{51} & \ding{51} & \ding{55} & \ding{55}& \ding{51} &\nn{23661} &\nn{537466}&Face forgery\\
			ForenSynths \cite{wang2020cnn} & \ding{51} & \ding{51} & \ding{55} & \ding{55} & \ding{51} & \nn{362000}&\nn{362000} &General forgery\\
			DE-FAKE \cite{sha2022defake} & \ding{55} & \ding{51} & \ding{55} & \ding{55} & \ding{55} &\nn{20000} &\nn{60000}&General forgery\\
			DiffusionForensics \cite{wang2023dire} & \ding{51} & \ding{51} & \ding{55} & \ding{55} & \ding{55} &\nn{134000} &\nn{481200}&General forgery\\
			DMimageDetection \cite{corvi2023detection} & \ding{51} & \ding{51}& \ding{55}& \ding{55}& \ding{55}&\nn{200000} &\nn{200000}&General forgery\\
			GenImage \cite{zhu2023genimage} & \ding{51} & \ding{51} & \ding{55} & \ding{55} & \ding{55} &\nn{1331167} &\nn{1350000}&General forgery\\
			RealHD (ours) & \ding{51} & \ding{51} & \ding{51} & \ding{51} & \ding{51} &\nn{204136}&\nn{531430}&General forgery\\
			\bottomrule
		\end{tabular}
	\end{table*}
	
	To summarize, our main contributions are as follows.
	\begin{itemize}
		\item \textbf{High quality \& large scale.} The RealHD dataset contains over \nn{730000} images, 88.75\% of which are stored in \texttt{PNG} format, and the remaining images are \texttt{JPEGs} with quality $\geq 90$.
		\item \textbf{Diversity \& realism.} RealHD leverages state-of-the-art generative models to produce synthetic images from over \nn{10000} carefully designed prompts, covering text-to-image generation, image refinement, image inpainting, and face swapping, resulting in diverse and realistic visual content.
		\item \textbf{Fine-grained annotations.} Each generated image is annotated with its generation method and category. For image inpainting, binary masks are provided to indicate the inpainted regions. For image refinement, each refined image is annotated with its corresponding source.
		\item \textbf{Simple \& effective baseline.} We introduce a lightweight detection approach based on noise entropy, which extracts NLM noise entropy from images for classification. Experimental results demonstrate significant performance improvement, establishing it as a baseline for future research.
	\end{itemize}

	\section{Related Work} 
    \subsection{AI-generated Image Datasets} 
    Several datasets for AI-generated image detection have been proposed, with early works primarily relying on GANs. For example, ForenSynths \cite{wang2020cnn} uses ProGAN \cite{karras2018progressive} to create its training set. With the rise of diffusion models \cite{ho2020denoising,song2021denoising,nichol2021improved,liu2022pseudo}, new datasets have emerged to broaden the scope of generative models and content diversity. Examples include DE-FAKE \cite{sha2022defake}, CIFAKE \cite{bird2024cifake}, DiffusionDB~\cite{wang2022diffusiondb}, DiffusionForensics \cite{wang2023dire}, DMimageDetection \cite{corvi2023detection} and GenImage \cite{zhu2023genimage}, which synthesize images using multiple generators. However, many datasets rely on short prompts (e.g., ``a photo of a cat'') and JPEG-compressed images, which limit the generalization.
\subsection{Detection of AI-Generated Images}

Early detection methods focused on spatial-domain features such as color \cite{mccloskey2018detecting_color}, saturation \cite{mccloskey2019detecting_saturation}, and co-occurrence \cite{nataraj2019detectinggangeneratedfake}, but they show limited generalization to newer models. Wang et al. \cite{wang2020cnn} demonstrate that classifiers trained solely on ProGAN can generalize surprisingly well to other GAN architectures. As an alternative, Frank et al. \cite{frank2020leveraging} analyze frequency-domain GAN artifacts from upsampling.
As diffusion models become prevalent, newer approaches emphasize broader generalization, including UnivFD~\cite{ojha2023towards}, which leverages CLIP-ViT features, and DIRE \cite{wang2023dire}, which measures reconstruction errors from a pretrained diffusion model. Other methods include Learned Noise Patterns (LNP) \cite{liu2022detecting} using frequency-domain noise, PatchCraft \cite{zhong2024patchcraft} exploiting texture inconsistencies, and AIDE~\cite{yan2025sanity} combining pixel statistics with semantic embeddings.

\section{RealHD Dataset Construction}
To detect increasingly sophisticated AI-generated images, we introduce RealHD, a comprehensive dataset comprising four key components: real images, generated images, detailed annotations, and a collection of generation prompts. Table \ref{tab:datasets_overview} provides an overview. In constructing the dataset, we adhere to the following core principles. 
\begin{enumerate}
	\item \textbf{High-quality real images.} All real images must be high-definition with lossless or minimally lossy compression.
	\item \textbf{Diverse technical coverage.} The generated images should include key tasks like image inpainting, text-to-image generation, face swapping, and image refinement.
	\item \textbf{Scenario coverage.} The generated images must cover both common real-world image themes and typical high-risk scenarios associated with generative models.
\end{enumerate}
\subsection{Collection of High-Definition Real Images}
The real images are collected using a multi-source acquisition strategy to ensure diversity.
To integrate global digital cultural heritage, we systematically collect high-resolution artworks from public repositories, prioritizing institutions with open-access licenses.
\begin{itemize} 
	\item \textbf{Western artworks.} Collected from the Metropolitan Museum of Art, the Louvre, the Rijksmuseum, the National Gallery (London), the Uffizi Gallery, the Prado Museum, Tate Britain, Google Arts \& Culture, and the Web Gallery of Art.
	 \item \textbf{Eastern artworks.} Including traditional Chinese ink paintings, ancient scrolls, and other forms of art from the Palace Museum, the National Palace Museum (Taipei), the National Museum of China, the Shanghai Museum, the Zhejiang Provincial Museum, the Nanjing Museum, the Hunan Provincial Museum, and the China National Digital Library.
\end{itemize}

All collected images are preserved as original high-resolution files in uncompressed \texttt{TIFF}, \texttt{RAW}, \texttt{PNG}, and high-quality \texttt{JPEG} formats. The collection spans over \nn{3300} years of visual culture, ranging from Shang Dynasty oracle bones (circa 1300 BC) to 20th-century artworks, and includes representative works from over \nn{1000} artists across different historical periods. It makes it one of the most temporally extensive visual cultural archives available to researchers.

In addition to artworks, we collect images in other categories from multiple sources. Professional photographs are captured using high-definition cameras, covering subjects such as animals and natural landscapes. In addition, we extract keyframes depicting animals and natural landscapes from publicly available BBC and CCTV video archives using motion-based frame differencing and keyframe selection algorithms. Public news photographs are collected via web crawlers from authoritative media platforms such as Xinhua News, BBC, and CNN, covering political, sports, cultural, and other sections. These images constitute a unique strength of our dataset, as AI-generated fake news can distort public opinion and influence political outcomes. Finally, high-quality portrait images are curated from well-known datasets such as Flickr-Faces-HQ~\cite{karras2021style} and BP4D~\cite{zhang2014bp4d}, ensuring they meet RealHD's strict quality standards.

\begin{figure*}[t]
	\centering
	\includegraphics[width=1.0\textwidth,keepaspectratio,trim=0mm 0mm 0mm 0mm, clip]{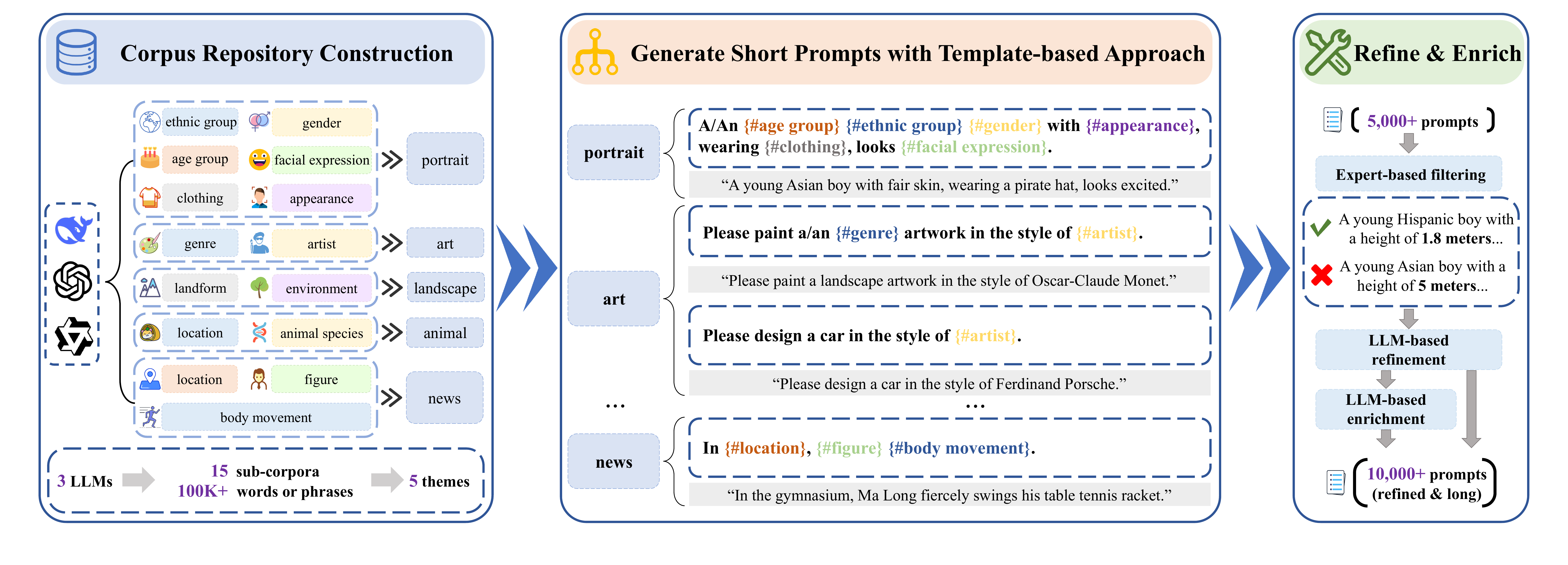}
	\Description{A figure illustrating the prompt construction pipeline. It shows three stages: building a corpus repository with 15 sub-corpora, generating short prompts using templates, and refining prompts with large language models to enhance fluency and diversity.}
	\caption{Overview of the prompt construction pipeline. The process consists of three key stages: (1) constructing a corpus repository containing 15 sub-corpora categorized by theme, (2) generating short prompts by instantiating predefined templates with expressions sampled from the sub-corpora, and (3) refining and enriching the prompts using large language models (LLMs) to improve fluency, semantic richness, and contextual diversity.}
	\label{fig:prompt_construction}
	\vspace{-2mm}
\end{figure*}

To standardize the images for training and testing, we perform resolution-based filtering on the collected images, discarding those with insufficient resolution. In addition, images in the \texttt{JPEG} format with a compression quality below 90 are excluded to ensure high visual fidelity. Subsequently, images in \texttt{TIFF} or \texttt{RAW} format are processed through demosaicing, tone mapping, and color space conversion before being saved as \texttt{PNG} with lossless compression and 8‑bit depth per channel to preserve high dynamic range and color fidelity.
Finally, as traditional East Asian long-scroll paintings exemplify extreme aspect ratios, all images with such ratios (e.g., width-to-height $\geq 2:1$) are cropped into square patches based on the short edge.
The RealHD dataset ultimately contains over \nn{200000} high-quality real-world images across five major categories: portrait, art, landscape, animal, and news, offering a comprehensive representation of the visual diversity of the real world.

Table \ref{tab:resolutions} shows the proportional distribution of images from all datasets over different resolution ranges, where \texttt{px} denotes pixels. The results show that RealHD images generally have higher resolutions, similar to those commonly found on social media.
   
    \begin{table}[h]
		\small
		\centering
		\caption{Proportional distribution of image resolutions.}
		\label{tab:resolutions}
		\scalebox{0.95}{
		\begin{tabular}{lcccc}
			\toprule
			\B{Dataset} & $10^3$--$10^4$ \texttt{px} & $10^4$--$10^5$ \texttt{px} & $10^5$--$10^6$ \texttt{px} & $\geq 10^6$ \texttt{px}\\
			\midrule
		GenImage &0.24\% & 29.99\% &62.87\% &6.90\%  \\
		DiffusionForensics & 0.01\% & 66.17\%	& 26.00\% & 7.82\% \\
		DMimageDetection & 0.00\% & 75.01\% & 24.99\% & 0.00\% \\
		RealHD (Ours) & 0.00\% & 0.46\% & 25.99\% & 73.55\% \\
			\bottomrule
		\end{tabular}}
		
	\end{table}
	
	\subsection{Text-to-Image Generation}
    One critical component of Text-to-Image (T2I) generation is the quality of prompts, as it directly influences the realism and diversity of the generated images. However, existing datasets typically use overly simple prompts,  such as ``a photo of \{category\}'', which lead to images lacking semantic richness and authenticity. To address this limitation, we carefully construct over \nn{10000} prompts, divided into refined and extended long prompts. The prompt construction process is detailed in Fig. \ref{fig:prompt_construction} and Algorithm \ref{alg:prompt}, as described below.

\begin{table*}
    \centering
    \small
    \caption{Comparison of demographic distributions in portrait prompts versus real-world population statistics.}
    \label{tab:demographic_comparison}
    \begin{tabular}{lcccccccc}
        \toprule
        \multirow{2}{*}{\B{Category}} 
        & \multicolumn{4}{c}{\B{Ethnic Groups by World Region}}
        & \multicolumn{4}{c}{\B{Age Group}} \\
       \cmidrule(lr){2-5}  \cmidrule(l){6-9} 
        & Eurasia & Africa & Americas & Oceania & $0$--$12$ & $13$--$18$ & $19$--$60$ & $> 60$ \\
        \midrule
        Rate in Portrait Prompts    & 47.5\% & 6.7\% & 37.6\% & 8.2\% & 11.3\% & 17.6\% & 60.3\% & 10.8\% \\
        Rate in Real World \cite{un_wpp2024} & 68.2\% & 18.3\% & 12.9\% & 0.6\% & 22.9\% & 10.3\% & 55.4\% & 11.4\% \\
        \bottomrule
    \end{tabular}
\end{table*}

	\begin{table}[h]
		\small
		\centering
		\caption{Category-wise statistics in the RealHD dataset.}
		\label{tab:realhd_category_stats}
		\begin{tabular}{{l@{\hspace{8pt}}c@{\hspace{8pt}}c@{\hspace{8pt}}c@{\hspace{8pt}}c@{\hspace{8pt}}c}}
			\toprule
			\B{Category} & Art & Landscape & Portrait & News & Animal \\
			\midrule
			Real Images & \nn{55045}    &\nn{15333}	&\nn{100001}	&\nn{22757}	&\nn{11000} \\
			Generated Images &\nn{144833}	&\nn{56021}	&\nn{299893}	&\nn{22079}	&\nn{8604}   \\
			\bottomrule
		\end{tabular}
		\vspace{-0.3cm}
	\end{table}

\begin{algorithm}[h]
		\caption{The Procedure of Prompt Generation}
		\label{alg:prompt}
		\begin{algorithmic}
			\STATE {\bfseries{Input:}} the major categories \{\emph{portrait}, \emph{art}, \emph{landscape}, \emph{animal}, \emph{news}\}, sub-corpora categories \{\emph{ethnic group}, \textellipsis{}, \emph{body movement}\}, and large language models (LLMs). 
			\STATE {\bfseries{Output:}} the final prompts used for generating images.
			\FOR{$c$ {\bfseries{in}} \{\emph{ethnic group}, \textellipsis{}, \emph{body movement}\}}
			\STATE $\text{raw\_output} \gets \text{LLMs.generate}(c, \text{instruction})$; \COMMENT{e.g., ``Please tell me the names of \nn{100} artists.''}
			\STATE $\text{sub-corpora}[c] \gets \text{HumanExpert.filter}(\text{raw\_output})$;
			\ENDFOR
			\STATE $\mathcal{P} \gets \emptyset$; \COMMENT{initialize an empty set to store generated prompts.}
			\FOR{$k$ {\bfseries{in}} \{\emph{portrait}, \emph{art}, \emph{landscape}, \emph{animal}, \emph{news}\}}
			\FOR{$i$ {\bfseries{in}} \{$1$, \textellipsis{}, $N$\}}
				\STATE $\mathcal{T}_{k,i} \gets \text{concat}(\tau_{i,1},\dots,\tau_{i,L_i})$, where $\tau_{i,1},\dots,\tau_{i,L_i}$ are predefined for the template, and $L_i$ denotes the length of $\mathcal{T}_{k,i}$;
				\STATE $\mathcal{P}_{k,i} \gets \text{fill\_template}(\mathcal{T}_{k,i}, \text{sub-corpora})$; \COMMENT{fill each slot $\phi^{\text{slot}}_{i,c}$ in $\mathcal{T}_{k,i}$ with an expression from $\text{sub-corpora}[c]$.}
				\STATE $\mathcal{P}.\text{add}(\mathcal{P}_{k,i})$;  \COMMENT{add the generated prompt $\mathcal{P}_{k,i}$ into $\mathcal{P}$.} 
			\ENDFOR
			\ENDFOR
			\STATE $\mathcal{P}_\text{filtered} \gets \text{HumanExpert.filter}(\mathcal{P});$
			\STATE $\mathcal{P}_\text{refined} \gets \text{LLMs.generate}(\mathcal{P}_\text{filtered}, \text{instruction})$; \COMMENT{e.g., ``Please improve the fluency and naturalness of this sentence.''}
			\STATE $\mathcal{P}_\text{long} \gets \text{LLMs.generate}(\mathcal{P}_\text{refined}, \text{instruction})$; \COMMENT{e.g., ``Please enrich this sentence with more content-specific details.''}	
			\STATE \textbf{return} $\mathcal{P}_\text{refined} \cup \mathcal{P}_\text{long}$;
		\end{algorithmic}
	\end{algorithm}

	\textbf{Corpus repository construction.} Based on the categories and styles of real images, we define 15 sub-corpora, which are categorized into five major groups, as shown in Fig. \ref{fig:prompt_construction}:
	\begin{itemize}
		\item \textbf{Portrait}: \emph{ethnic group}, \emph{gender}, \emph{age group}, \emph{facial expression}, \emph{clothing}, \emph{appearance}
		\item \textbf{Art}: \emph{genre}, \emph{artist}
		\item \textbf{Landscape}: \emph{landform}, \emph{environment}
		\item \textbf{Animal}: \emph{location}, \emph{animal species}
		\item \textbf{News}: \emph{location}, \emph{figure}, \emph{body movement}
	\end{itemize}
	Leveraging pre-trained LLMs---including DeepSeek-V2 \cite{deepseek2024v2}, GPT-4o \cite{openai2024gpt4osystemcard}, and Qwen-72B \cite{qwen}---we generate multiple expressions for each sub-corpus, with each expression corresponding to a domain-specific word or phrase. While the majority of sub-corpora contain thousands of expressions, a few---such as the \emph{gender} sub-corpus---contain only a limited set.
 	These 15 sub-corpora collectively form the \emph{corpus repository}. Subsequently, human experts filter the generated expressions to ensure high relevance to their corresponding sub-corpora. After filtering, the final corpus repository consists of over \nn{100000} expressions. 
	
	\textbf{Templates of prompts.} Once the corpus repository is constructed, we define structured generation templates for each of the five categories---\emph{portrait}, \emph{art}, \emph{landscape}, \emph{animal}, and \emph{news}---to generate semantically coherent and contextually appropriate prompts. The set of templates for each category is formally defined as:
	\begin{equation}
		\mathcal{T}_{k} \coloneqq \left\{ \mathcal{T}_{k,i} : \mathcal{T}_{k,i} \coloneqq \text{concat}(\tau_{i,1},\dots,\tau_{i,L_i}), \; 1 \leq i \leq N \right\},
	\end{equation}
	where $k$ is a placeholder for the category selected from \emph{portrait}, \emph{art}, \emph{landscape}, \emph{animal}, and \emph{news}, $\tau_{i,j} \in \left\{ \phi^{\text{fix}}_{i,j}, \phi^{\text{slot}}_{i,c} \right\}$ denotes the $j$-th expression of the $i$-th template, $L_i$ denotes the length (in tokens) of the $i$-th template, and $N$ denotes the total number of templates for the category $k$. Specifically, $\phi^{\text{fix}}_{i,j}$ represents the $j$-th predefined fixed expression for the $i$-th template (e.g., \textit{Please}, \textit{in the style of}), while $\phi^{\text{slot}}_{i,c}$ denotes the $i$-th template's slot that can be filled with a randomly selected expression from the sub-corpus $c$, and $c$ is the name of a sub-corpus such as \emph{artist} or \emph{genre} (see the previous paragraph).
	Let us take a template from the \emph{art} category as an example. We define the $i$-th template as $\mathcal{T}_{\emph{art},i} \coloneqq \text{concat}(\phi^{\text{fix}}_{i,1}, \phi^{\text{slot}}_{i,\emph{genre}}, \phi^{\text{fix}}_{i,3}, \phi^{\text{slot}}_{i,\emph{artist}})$, where $\phi^{\text{fix}}_{i,1}$ denotes ``\textit{Please paint a}'', $\phi^{\text{slot}}_{i,\emph{genre}}$ is a randomly selected expression from the \emph{genre} sub-corpus (e.g., ``\textit{landscape}''), $\phi^{\text{fix}}_{i,3}$ denotes ``\textit{artwork in the style of}'', and $\phi^{\text{slot}}_{i,\emph{artist}}$ is a randomly selected expression from the \emph{artist} sub-corpus (e.g., ``\textit{Oscar-Claude Monet}''). 
	Therefore, the final sentence becomes: ``\textit{Please paint a landscape artwork in the style of Oscar-Claude Monet.}'' Using this method, we construct \nn{5000} short prompts, which will be refined and enriched.
	
	\textbf{Refine and enrich short prompts.}  After obtaining \nn{5000} short prompts, we manually review them to identify semantically conflicting prompts, including logical contradictions and physical impossibilities. Such prompts are filtered out (e.g., ``a five-meter-tall boy''). The remaining prompts are then refined using the LLMs to improve fluency and clarity, thereby producing a high-quality subset. Subsequently, the LLMs enrich the refined prompts by adding imaginative and contextually rich details, thereby producing extended versions (called long prompts). After completing the entire processing pipeline, we generate more than \nn{10000} high-quality prompts, with short and long prompts averaging 19 and 30 tokens, respectively.
    In addition, to promote diversity in generated images, we emphasize the richness of the prompt content. For example, the prompts in the \emph{portrait} category span all age groups, from infants to the elderly, as shown in Table~\ref{tab:demographic_comparison}. To ensure ethnic diversity, the prompts cover ethnic groups across four major world regions, from Africa to the Americas. This distribution is broadly comparable to the United Nations' World Population Prospects 2024 \cite{un_wpp2024}, with adjustments to enhance diversity.

    \textbf{Generate images with crafted prompts.} To cover the main technical branches of generative models in practical application scenarios, we select the 8 most popular models to generate images, including Flux and several variants of Stable Diffusion (v1.5, v2.1, v3.0, v3.5, XL-base, XL-Turbo, and Cascade) \cite{arombach2022high,podell2024sdxl}. Table \ref{tab:realhd_category_stats} shows the total image count for each category in our dataset, highlighting the diversity across the image categories.
	\subsection{Image Inpainting}
	Image inpainting refers to the process of filling in missing or corrupted regions of an image in a visually plausible way. The goal is to restore the image such that the completed content is coherent in terms of texture, structure, and semantics. We use image inpainting to generate synthetic images, which are then incorporated into the RealHD dataset.
    Our algorithm mimics the behavior of manually erasing parts of an image and regenerating the missing regions. For each real image $\mathbf{x} \in \mathbb{R}^{H \times W \times 3}$, where $H$ and $W$ denote the height and width of $\mathbf{x}$, we define a binary mask $M \in \{0,1\}^{H \times W}$, where the white region ($M(x, y) = 1$) corresponds to the area to be inpainted, and the black region ($M(x, y) = 0$) preserves the original content. The white regions are generated by simulating the motion of a circular brush, which starts at a random position within the image and moves back and forth, mimicking the action of manually erasing with a handheld eraser. The original image $\mathbf{x}$ and mask $M$ are then fed into Stable Diffusion XL Inpainting \cite{podell2024sdxl} and Stable Diffusion 2 Inpainting \cite{arombach2022high} to perform image inpainting.
    \subsection{Image Refinement}
    Image refinement is a technique used to enhance the quality of generated images by refining coarse or degraded inputs. It improves visual fidelity by adding fine-grained details, removing artifacts, and making the output appear more natural and realistic.
    In this study, we apply image refinement to images affected by various degradations such as noise, blur, and low resolution, resulting in a high-quality refined image set.
    For instance, when Stable Diffusion creates an image with a resolution of $512 \times 512$, the Stable Diffusion XL Refiner (SDXL Refiner) \cite{podell2024sdxl} enhances its resolution to $1024 \times 1024$ while improving its overall quality. Additionally, for images that are noisy or blurred, the SDXL Refiner reduces noise and restores fine details, producing refined images with improved visual realism.

    \subsection{Face Swapping}
    Face swapping is a computer vision technique that replaces the facial identity of a source person in an image or video with that of a target person, while preserving the source's original facial expression, pose, and lighting conditions to create a realistic and seamless output.
   	To support this task, we construct two facial image sets: a source set $S$ and a target set $T$, each containing distinct identities, such that $S \cap T = \emptyset$. The source set $S$ consists of 201 facial images with unique identities, while the target set $T$ contains 200 facial images, each with a different identity. For each source image $\mathbf{x}_\text{source} \in S$, we apply Face-Adapter \cite{han2024face} to generate face-swapped images with each target image in $T$. This procedure produces a total of \nn{40200} ($201\times200$) face-swapped images by swapping each source face with all target faces. Finally, we manually filter these images, retaining \nn{40139} valid images and removing \nn{61} invalid ones.
    
    \section{Image Noise Entropy for AI-Generated Image Detection}    
	Recent studies \cite{liu2022detecting,wang2023dire} have shown that the low-level noise distribution can effectively distinguish real images from generated ones. Such low-level noise is often manifested in the high-frequency components of an image. However, traditional deep neural networks (DNNs) tend to focus on low-frequency features during training, while high-frequency features are more difficult to capture \cite{xu2019training,rahaman2019spectral}. On the other hand, real images captured by cameras inherently contain sensor noise originating from the image sensor, which differs from that in generated images. Therefore, we propose leveraging image noise entropy as a statistical descriptor of the uncertainty inherent in the noise distribution, serving as a discriminative feature for distinguishing real and generated images. Steps are as follows.

\textbf{Image noise extraction}. We adopt NLM \cite{buades2011NLM} to extract image noise $\eta$. Let $\mathbf{x}\in\mathbb R^{H\times W\times C}$ denote the original image, and let $\hat{\mathbf{x}}$ be the NLM-denoised image. Each pixel in $\hat{\mathbf{x}}$ is computed as
\begin{equation}
    \hat{\mathbf{x}}_i \coloneqq \sum_{j \in \Omega} w(i,j) \cdot \mathbf{x}_j,
\end{equation}
where $\Omega$ denotes the predefined non-local search window centered at pixel $i$, and $w(i,j)$ represents the similarity weight between pixels $i$ and $j$, satisfying the normalization constraint $\sum_{j \in \Omega} w(i,j) = 1$. $w(i,j)$ is defined as a Gaussian-weighted similarity measure based on the Euclidean distance between patches:
\begin{equation}
    w(i,j)\coloneqq \frac{1}{Z(i)} \cdot \exp{\left(-\frac{\| P(i) - P(j) \|_2^2}{h^2}\right)},
\end{equation}
where $P(i)$ and $P(j)$ denote the square patches centered at the $i$-th and $j$-th pixels, respectively. $\|P(i) - P(j)\|_2^2$ represents the squared Euclidean distance between the two patches. The parameter $h$ controls the filtering strength by regulating the sensitivity to patch similarity. $Z(i)$ is a normalization factor that ensures the weights sum to $1$.
Then, we calculate the residual noise map as
\begin{equation}
\eta \coloneqq \mathbf{x} - \hat{\mathbf{x}}.
\end{equation}

\textbf{Image noise entropy computation.} This noise map $\eta$ is further resized to $3\times 1024 \times 1024$ using bicubic interpolation, obtaining a tensor $\hat{\eta}\in\mathbb{R}^{3\times1024\times 1024}$, where $3$ and $1024$ correspond to the channel and spatial dimensions, respectively. Inspired by local Shannon entropy \cite{wu2013local}, we compute the Shannon entropy for each $\frac{1024}{n} \times \frac{1024}{n}$ block of $\hat{\eta}$ to obtain a size-reduced tensor $\mathcal{H} \in \mathbb{R}^{3\times n\times n}$, where $n$ represents the spatial resolution after block-wise Shannon entropy computation. The detailed steps are as follows.

\paragraph{Step 1: Block Partitioning}
The resized noise map $\hat{\eta} \in \mathbb{R}^{3\times 1024 \times 1024}$ is uniformly divided into $n \times n$ non-overlapping local blocks. Each block is denoted as $B_{i,j}^c$, where $c$ is the channel index, $i$ and $j$ denote the row and column indices of the block. The subsequent computation is independently performed on each channel.

\paragraph{Step 2: Discretized Shannon Entropy Computation}
For each block $B_{i,j}^c$, the discretized Shannon entropy $\mathcal{H}_c(i, j)$ is calculated as:
\begin{align}
\mathcal{H}_c(i, j) &= - \sum_{k} p_k \cdot \log p_k, \\
p_k &\coloneqq \frac{1}{N}\sum_{x\in B_{i,j}^c} \mathbb{I}[x \in I_k],
\end{align}
where:
\begin{itemize}
    \item $x$ denotes a noise value of a pixel within the block $B_{i,j}^c$.
    \item $I_k$ denotes the $k$-th quantization interval, obtained by uniformly dividing the range of noise values in $\hat{\eta}$ into a fixed number of bins.
    \item $\mathbb{I}[\cdot]$ is the indicator function.
    \item $N$ is the number of pixels in each block.
\end{itemize}

\paragraph{Step 3: Channel-wise Tensor Concatenation}
$\mathcal{H}_c(i,j)$ from all channels are stacked to form the final entropy tensor $\mathcal{H} \in \mathbb{R}^{3 \times n \times n}$.

\begin{figure}[h]
	\centering
	\begin{subfigure}{0.235\textwidth} 
		\centering
		\includegraphics[width=\linewidth]{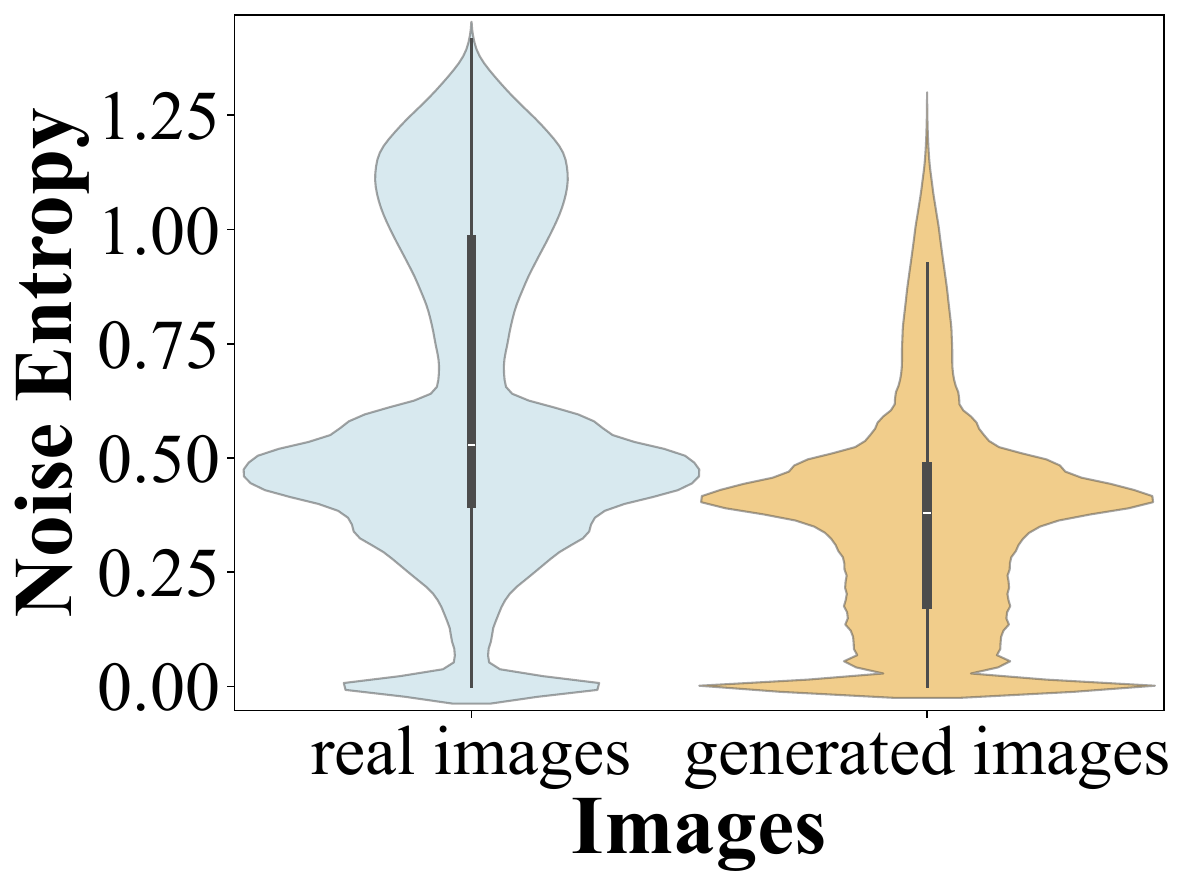}
		\Description{Violin plot of noise entropy distributions for RealHD dataset}
		\caption{The RealHD dataset}
		\label{subfig:RealHD_violin_plot}
	\end{subfigure}
	\begin{subfigure}{0.235\textwidth} 
		\centering
		\includegraphics[width=\linewidth]{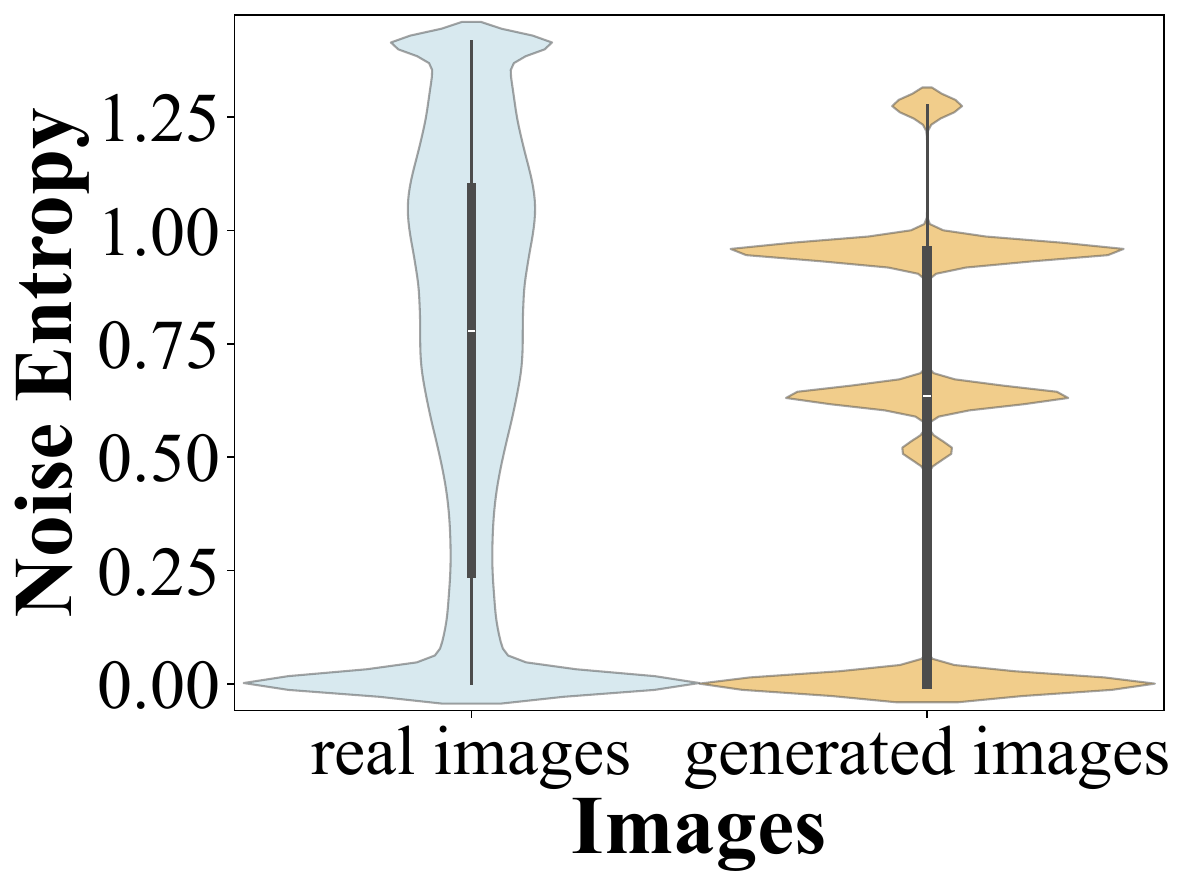}
		\Description{Violin plot of noise entropy distributions for GenImage dataset}
		\caption{The GenImage dataset}
		\label{subfig:GenImage_violin_plot}
	\end{subfigure}
	\caption{Comparison of noise entropy distributions between real and generated images in the RealHD and GenImage datasets (based on 1,000 sampled images). These differences motivate a lightweight detection method based on image noise entropy, which improves model generalizability.}
	\label{fig:image_noise_entropy}
\end{figure}

This entropy tensor encodes statistical features that discriminate between real and generated images, as real images typically contain sensor noise introduced during acquisition that differs from that in generated images. As illustrated by the violin plot (Fig. \ref{fig:image_noise_entropy}), the image noise entropy distributions of real and generated images exhibit clear discrepancies, providing a reliable cue for detection across various generative models.
    
	\section{Experiment}
	We first evaluate the RealHD dataset by comparing it with other mainstream datasets, and then assess the proposed detection method.
    \subsection{Dataset Evaluation}
    \textbf{Compared datasets.} We compare three representative datasets:
    \begin{itemize}
    	\item \textbf{GenImage} \cite{zhu2023genimage}: the largest open-source benchmark in terms of data scale and category diversity;
    	\item \textbf{DiffusionForensics} \cite{wang2023dire}: a comprehensive dataset for evaluating detection performance on diffusion-generated images;
    	\item \textbf{DMimageDetection} \cite{corvi2023detection}: it covers multiple generative architectures, including diffusion models and GANs, and incorporates compression and resizing to simulate real-world post-processing as seen on social media platforms.
    \end{itemize}

    \begin{table}
        \small
        \centering
        \caption{Aesthetic scores assessed by SAAN, where DF and DM represent DiffusionForensics and DMimageDetection.}
        \label{tab:SAAN}
        \begin{tabular}{lcccc}
            \toprule
            \multirow{2}{*}{\B{Method}} & \multicolumn{4}{c}{\B{Dataset}} \\
            \cmidrule(l){2-5}
            & GenImage
            & DF
            & DM
            & RealHD (Ours)
            \\
            \midrule
            SAAN & 3.889 & 3.823 & 3.892 & \B{4.016} \\
            \bottomrule
        \end{tabular}
\end{table}

\noindent\textbf{Aesthetic analysis.}
In real-world scenarios, users often have high aesthetic demands for generated images when using generative models. To this end, we evaluate the aesthetic quality of the entire test set from each of the three compared datasets and RealHD using the Style-specific Art Assessment Network (SAAN)~\cite{yi2023SAAN}. SAAN assesses aesthetics from multiple perspectives by extracting style features via adaptive instance normalization, general aesthetic features through self-supervised pre-training, and spatial layout information using non-local blocks. As shown in Table \ref{tab:SAAN}, the RealHD dataset achieves the highest average aesthetic score among all datasets, demonstrating its superior visual quality and making it particularly valuable for applications that demand high aesthetic standards.

\begin{table}[ht]
    \centering
     \small
    \caption{Performance metrics across different models and datasets, where lower values suggest more challenging datasets (lower is better). ACC denotes Accuracy, and AUC refers to the Area Under the ROC Curve.}
    \label{tab:performance}
    \vspace{-2mm}
    
    \begin{tabular}{llccc}
        \toprule
        \B{Method} & \B{Dataset} & \B{ACC} & \B{F1 score} & \B{AUC}\\
        \midrule
        
        \multirow{4}{*}{ResNet-50} 
        & GenImage          & 94.29\% & 94.03\% & 0.9904 \\
        & DiffusionForensics & \underline{88.41\%} & \underline{92.90\%} & \B{0.9186} \\
        & DMimageDetection       & 96.03\% & 95.91\% & 0.9966 \\
        & RealHD (Ours)     & \B{83.29\%} & \B{86.99\%} & \underline{0.9277} \\
        \hline

        \multirow{4}{*}{Xception}
        & GenImage          & 99.00\% & 99.00\% & 0.9996 \\
        & DiffusionForensics & \underline{95.71\%} & \underline{97.40\%} & \underline{0.9969} \\
        & DMimageDetection       & 99.78\% & 99.78\% & 0.9999 \\
        & RealHD (Ours)     & \B{85.21\%} &\B{90.16\%} & \B{0.9515} \\
        \hline
        
        \multirow{4}{*}{EfficientFormer}
        & GenImage          & 98.71\% & 98.71\% & 0.9988 \\
        & DiffusionForensics & \underline{96.04\%} & \underline{97.62\%} & \underline{0.9843} \\
        & DMimageDetection       & 99.33\% & 99.34\% & 0.9997 \\
        & RealHD (Ours)     & \B{81.66\%} & \B{88.03\%} & \B{0.9535} \\
        \hline
        
        \multirow{4}{*}{LNP}
        & GenImage          & \underline{97.68\%} & \B{97.23\%} & \underline{0.9974} \\
        & DiffusionForensics & 97.91\% & 98.47\% & 0.9980 \\
        & DMimageDetection       & 98.58\% & 98.59\% & 0.9989 \\
        & RealHD (Ours)     & \B{96.50\%} & \underline{97.53\%} & \B{0.9950} \\
        \bottomrule
    \end{tabular}
    
\end{table}

\noindent\textbf{Comparison of datasets through different detectors.}
In real-world scenarios, generated images are often highly diverse and complex. To accurately capture the challenges presented by such scenarios, a dataset must possess sufficient complexity. This complexity typically results in lower detection performance across various models, suggesting that the dataset is more challenging and thus more effective for evaluating the generalization capabilities of detection methods. To this end, we conduct experiments on three backbone architectures, i.e., ResNet-50 \cite{he2016deep}, Xception \cite{chollet2017xception}, and EfficientFormer \cite{li2022efficientformer}. Additionally, we also evaluate LNP \cite{liu2022detecting}, a detection method based on the discriminative noise patterns.

All models are trained for 15 epochs using the Adam optimizer with an initial learning rate of $0.00125$ and a batch size of $64$. The images are resized to $224 \times 224$ and augmented with random cropping and horizontal flipping. Our RealHD dataset is split into 80\% for training and 20\% for testing. To address class imbalance and provide a comprehensive evaluation of model performance, we report classification accuracy, F1 score, and AUC.
As shown in Table~\ref{tab:performance}, the RealHD dataset yields lower accuracy, F1 score, and AUC across various detection models compared with the other three datasets, indicating that it is more complex and challenging. This can be attributed to the greater diversity of generative methods and the higher visual quality of the images included in the RealHD dataset.

\begin{table}[htbp]
    \centering
    \caption{Cross-domain detection accuracy, where DF and DM denote DiffusionForensics and DMimageDetection.}
    \label{tab:cross-domain}
    \vspace{-2mm}
    \scalebox{0.9}{
    \begin{tabular}{@{}l@{\hspace{6pt}}l@{\hspace{4pt}}c@{\hspace{4pt}}c@{\hspace{4pt}}c@{\hspace{4pt}}c@{\hspace{6pt}}c@{}}
        \toprule
        \multirow{2}{*}{\B{Method}} & \multirow{2}{*}{\B{Training set}} & \multicolumn{5}{c}{\B{Test set (\%)}} \\
        \cmidrule(l){3-7}
        & & \multicolumn{1}{c}{GenImage} & \multicolumn{1}{c}{DF} & \multicolumn{1}{c}{DM} & \multicolumn{1}{c}{RealHD} & \multicolumn{1}{c}{Avg.} \\
        \midrule
        \multirow{4}{*}{ResNet-50} 
        & GenImage & -- & \underline{71.22} & \B{84.30} & \underline{49.18} & \underline{68.23} \\
        & DF & 58.37 & -- & 59.21 & \B{66.53} & 61.37 \\
        & DM & \underline{69.27} & 56.94 & -- & 36.60 & 54.27 \\
        & RealHD & \B{75.92} & \B{78.19} & \underline{63.40} & -- & \B{72.50} \\
        \hline
        \multirow{4}{*}{Xception}
        & GenImage & -- & \underline{77.60} & \B{93.36} & \underline{58.56} & \underline{76.51} \\
        & DF & 67.40 & -- & 73.78 & \B{71.71} & 70.96 \\
        & DM & \underline{69.67} & 59.34 & -- & 32.31 & 53.77 \\
        & RealHD & \B{73.52} & \B{78.41} & \underline{81.30} & -- & \B{77.74} \\
        \bottomrule
    \end{tabular}}
\end{table}
    
    \noindent\textbf{Cross-domain evaluation.} To assess the quality of each dataset, we perform cross-domain evaluations by training models on one dataset and testing them on another. This simulates real-world scenarios with out-of-distribution data, where better performance indicates stronger generalization.
    As shown in Table \ref{tab:cross-domain}, RealHD-trained models achieve the best cross-domain performance, with Xception reaching the highest average accuracy. These results confirm that RealHD is an effective dataset for enabling superior model generalization in cross-domain settings.
  
    \noindent\textbf{Generalization evaluation on the Chameleon dataset.}
    To assess the generalization capability of various datasets, we train models on each dataset and evaluate their performance on Chameleon~\cite{yan2025sanity}. Chameleon contains AI-generated images from popular AI-painting communities (i.e., ArtStation, Civitai, and Liblib), created using tools like Midjourney, DALL-E 3, and LoRA modules based on Stable Diffusion. All images were manually screened to include only those that passed a human perception ``Turing Test'', where annotators misclassified them as real.
    With perceptual realism, diversity, and high resolution, Chameleon serves as a challenging benchmark, enabling us to determine whether models trained on RealHD generalize better than those trained on other datasets.
    
    To comprehensively evaluate performance on Chameleon, we evaluate three additional methods: LNP \cite{liu2022detecting}, Neighboring Pixel Relationships (NPR) \cite{tan2024NPR}, and FatFormer \cite{liu2024fatformer}.
    As shown in Table \ref{tab:Chameleon}, models trained on RealHD consistently outperform those trained on other datasets, achieving significant improvements in both accuracy and AUC.
    These results highlight the superior generalization of our dataset and its practical applicability in real-world scenarios. 

\begin{table}[ht]
		\centering
		\caption{Experimental results on the Chameleon dataset.}
		\label{tab:Chameleon}

		\begin{tabular}{llcc}
			\toprule
			\B{Method} & \B{Training Dataset} & \B{ACC} & \B{AUC} \\
			\midrule
			
			\multirow{4}{*}{ResNet-50}
			& GenImage  & \underline{58.61\%} & \underline{0.5389} \\
			& DiffusionForensics     & 55.44\% & 0.5341 \\
			& DMimageDetection        & 56.65\% & 0.4121 \\
			& RealHD (Ours)      & \B{60.45\%} & \B{0.6549} \\
			\hline
			
			\multirow{4}{*}{Xception}
			& GenImage & 59.36\% & 0.4915 \\
			& DiffusionForensics      & \underline{60.87\%} & \underline{0.6549} \\
			& DMimageDetection        & 57.33\% & 0.3291 \\
			& RealHD (Ours)      & \B{66.25\%} & \B{0.7202} \\
			\hline
			
			\multirow{4}{*}{EfficientFormer}
			& GenImage & \underline{59.45\%} & \underline{0.7692} \\
			& DiffusionForensics      & 57.61\% & 0.6090 \\
			& DMimageDetection        & 57.31\% & 0.4796 \\
			& RealHD (Ours)      & \B{71.47\%} & \B{0.7811} \\
			\hline
			
			\multirow{4}{*}{LNP}
			& GenImage & 55.44\% & 0.4815 \\
			& DiffusionForensics      & 49.94\% & \underline{0.5375} \\
			& DMimageDetection        & \B{57.09\%} & 0.4722 \\
			& RealHD (Ours)      & \underline{55.99\%} & \B{0.5793} \\
			\hline
			 \multirow{4}{*}{NPR}
			& GenImage & 59.19\% & 0.5795 \\
			& DiffusionForensics      & \underline{59.60\%} & 0.5637 \\
			& DMimageDetection        & 57.40\% & \underline{0.6196} \\
			& RealHD (Ours)      & \B{59.70\%} & \B{0.6210} \\
			\hline
			
			\multirow{4}{*}{FatFormer}
		& GenImage & 42.91\% & 0.4970 \\
		& DiffusionForensics      & 48.92\% & 0.4207 \\
		& DMimageDetection        & \underline{51.92\%} & \underline{0.5426} \\
		& RealHD (Ours)      & \B{59.60\%} & \B{0.6465}\\
			\bottomrule
		\end{tabular}
		
	\end{table}
    
	\subsection{Detection Method Evaluation}
    \noindent\textbf{Performance comparison with and without noise entropy.}
    In this section, we evaluate the proposed image noise entropy method (NE) on the RealHD dataset.
    As shown in Table~\ref{tab:gen-test}, NE significantly improves detection performance across all models, outperforming models of the same architecture when trained directly on original images or extracted noise (denoted as ``Noise'' in Table \ref{tab:gen-test}). Furthermore, compared to LNP \cite{liu2022detecting}, which utilizes learned noise patterns, our NE method achieves better overall performance.
    
    \begin{table}[ht]
    \centering
    \caption{Comparison of different methods on the RealHD dataset, where ACC stands for accuracy.}
    \vspace{-2mm}
    \label{tab:gen-test}  
    \begin{tabular}{lccc}  
        \toprule
        \B{Method}         & \B{ACC} & \B{F1 score} & \B{AUC} \\
        \midrule
        ResNet-50               & 83.29\%       & 86.99\%            & 0.9277         \\
        Xception                & 85.21\%       & 90.16\%            & 0.9515         \\
        EfficientFormer         & 81.66\%       & 88.03\%            & 0.9535         \\
        LNP                     & 96.50\%       & 97.53\%            & 0.9950         \\
        \hline
        ResNet-50 + Noise    &93.95\%       & 93.56\%            & 0.9924          \\
        Xception + Noise      &\underline{98.40}\% & 98.38\% & \underline{0.9987} \\
        EfficientFormer + Noise    & 96.86\%       & 96.84\%  & 0.9957\\
        \hline
        ResNet-50 + NE    &97.94\%       & 98.48\%            & 0.9978          \\
        Xception + NE      &\B{98.76\%} & \B{99.09\%} & \B{0.9991} \\
        EfficientFormer + NE    & 98.29\%       & \underline{98.74\%}  & 0.9979\\
        \bottomrule
    \end{tabular}
\end{table}

     \noindent\textbf{Robustness against JPEG compression.}
    Images shared online are often compressed to reduce file size, which can significantly degrade classifier performance. To assess robustness under such conditions, we evaluate model performance under JPEG compression with quality factors $Q=90$, $75$, and $50$.
    As shown in Table~\ref{tab:JPEG}, while all models experience performance degradation with increased compression, models with NE---especially EfficientFormer---tend to maintain higher AUCs compared to the baselines, indicating improved robustness.
    This robustness may stem from the persistence of structured sensor noise in real images after compression, whereas generated images exhibit distinctive noise patterns, allowing NE to remain effective in distinguishing them. 
    
\begin{table}[ht]
    \centering
    \small
    \caption{Performance under JPEG Compression on RealHD.}
    \label{tab:JPEG}
    \vspace{-2mm}
    \begin{tabular}{lccc}  
        \toprule
        \multirow{2}{*}{\B{Method}} & \multicolumn{3}{c}{\B{AUC}} \\
        \cmidrule(l){2-4}
        & JPEG $Q=90$  & JPEG $Q=75$  & JPEG $Q=50$   \\
        \midrule
        ResNet-50       & 0.6747 & 0.5807 & 0.4991  \\
        Xception        & \underline{0.7328} & 0.6209 & 0.5370 \\
        EfficientFormer  & 0.7010 & 0.6173 & 0.5829 \\
        \hline
        ResNet-50 + NE    & 0.5670 & 0.5078 & 0.4428 \\
        Xception + NE       & 0.7071 & \underline{0.6256} & \underline{0.5875} \\
		EfficientFormer + NE & \B{0.7680} &\B{0.6735} &\B{0.5900} \\
        \bottomrule
    \end{tabular}
   \vspace{-0.35cm}
\end{table}
    
	\section{Conclusion}
    In this paper, we present RealHD, a high-quality dataset of over \nn{730000} images across multiple categories, addressing the limitations of existing datasets and enabling more generalizable detection models. RealHD offers diverse, well-annotated images with rich metadata, serving as a strong benchmark for detecting AI-generated images. We also propose a lightweight detection method based on image noise entropy, which achieves competitive performance and serves as a robust baseline in this field. Extensive experiments confirm the effectiveness of both our dataset and the proposed method, establishing a solid foundation for future research.

\begin{acks}
	This work was supported by Key R\&D Program of Zhejiang Province under Grant No. 2024C01164, and by Zhejiang Provincial Natural Science Foundation of China under Grant No. LMS25F020005.
\end{acks}

\bibliographystyle{ACM-Reference-Format}
\bibliography{main}

@InProceedings{arombach2022high,
	author={Rombach, Robin and Blattmann, Andreas and Lorenz, Dominik and Esser, Patrick and Ommer, Björn},
	booktitle={Proceedings of the IEEE/CVF Conference on Computer Vision and Pattern Recognition}, 
	title={High-Resolution Image Synthesis with Latent Diffusion Models}, 
	year={2022},
	pages={10674-10685},
	//keywords={Training;Visualization;Image synthesis;Computational modeling;Noise reduction;Superresolution;Process control;Image and video synthesis and generation},
	doi={10.1109/CVPR52688.2022.01042}}

@inproceedings{podell2024sdxl,
	title={{SDXL}: Improving Latent Diffusion Models for High-Resolution Image Synthesis},
	author={Dustin Podell and Zion English and Kyle Lacey and Andreas Blattmann and Tim Dockhorn and Jonas M{\"u}ller and Joe Penna and Robin Rombach},
	booktitle={International Conference on Learning Representations},
	year={2024},
	url={https://openreview.net/forum?id=di52zR8xgf}
}

@article{ferreira2020review,
	title = {A Review of Digital Image Forensics},
	journal = {Computers \& Electrical Engineering},
	volume = {85},
	pages = {106685},
	year = {2020},
	issn = {0045-7906},
	doi = {10.1016/j.compeleceng.2020.106685},
	author = {William D. Ferreira and Cristiane B.R. Ferreira and Gelson {da Cruz Júnior} and Fabrizzio Soares},
}

@inproceedings{xu2023combating,
	author = {Xu, Danni and Fan, Shaojing and Kankanhalli, Mohan},
	title = {Combating Misinformation in the Era of Generative AI Models},
	year = {2023},
	isbn = {9798400701085},
	publisher = {Association for Computing Machinery},
	address = {New York, NY, USA},
	url = {https://doi.org/10.1145/3581783.3612704},
	doi = {10.1145/3581783.3612704},
	booktitle = {Proceedings of the 31st ACM International Conference on Multimedia},
	pages = {9291–9298},
	//numpages = {8},
	//keywords = {aigc, generative ai models, misinformation detection, multimodal},
	//location = {Ottawa ON, Canada},
	series = {MM '23}
}

@preprint{ren2024copyright,
	author         = {Jie Ren and Han Xu and Pengfei He and Yingqian Cui and Shenglai Zeng and Jiankun Zhang and Hongzhi Wen and Jiayuan Ding and Pei Huang and Lingjuan Lyu and Hui Liu and Yi Chang and Jiliang Tang},
	title          = {Copyright Protection in Generative AI: A Technical Perspective},
	year           = {2024},
	archivePrefix  = {arXiv},
	eprint         = {2402.02333},
	doi            = {10.48550/arXiv.2402.02333}
}

@inproceedings{li2018UADFV,
	author={Li, Yuezun and Chang, Ming-Ching and Lyu, Siwei},
	booktitle={IEEE International Workshop on Information Forensics and Security}, 
	title={In Ictu Oculi: Exposing AI Created Fake Videos by Detecting Eye Blinking}, 
	year={2018},	
	pages={1-7},
	//keywords={Face;Gallium nitride;Training;Generators;Biological neural networks},
	doi={10.1109/WIFS.2018.8630787}}

@InProceedings{he2021forgerynet,
	author={He, Yinan and Gan, Bei and Chen, Siyu and Zhou, Yichun and Yin, Guojun and Song, Luchuan and Sheng, Lu and Shao, Jing and Liu, Ziwei},
	booktitle={Proceedings of the IEEE/CVF Conference on Computer Vision and Pattern Recognition}, 
	title={{ForgeryNet}: A Versatile Benchmark for Comprehensive Forgery Analysis}, 
	year={2021},
	pages={4358-4367},
	//keywords={Location awareness;Image segmentation;Technological innovation;Annotations;Face recognition;Perturbation methods;Benchmark testing},
	doi={10.1109/CVPR46437.2021.00434}}

@inproceedings{sha2022defake,
	author = {Sha, Zeyang and Li, Zheng and Yu, Ning and Zhang, Yang},
	title = {{DE-FAKE}: Detection and Attribution of Fake Images Generated by Text-to-Image Generation Models},
	year = {2023},
	isbn = {9798400700507},
	publisher = {Association for Computing Machinery},
	address = {New York, NY, USA},
	//url = {https://doi.org/10.1145/3576915.3616588},
	doi = {10.1145/3576915.3616588},
	booktitle = {Proceedings of the ACM SIGSAC Conference on Computer and Communications Security},
	pages = {3418–3432},
	numpages = {15},
	//keywords = {fake image detection, prompt analysis, text-to-image models},
	//location = {Copenhagen, Denmark},
	series = {CCS '23}
}

@ARTICLE{bird2024cifake,
	author={Bird, Jordan J. and Lotfi, Ahmad},
	journal={IEEE Access}, 
	title={{CIFAKE}: Image Classification and Explainable Identification of AI-Generated Synthetic Images}, 
	year={2024},
	volume={12},
	pages={15642--15650},
	doi={10.1109/ACCESS.2024.3356122}}

@inproceedings{wang2023dire,
	author={Wang, Zhendong and Bao, Jianmin and Zhou, Wengang and Wang, Weilun and Hu, Hezhen and Chen, Hong and Li, Houqiang},
	booktitle={Proceedings of the IEEE/CVF International Conference on Computer Vision}, 
	title={{DIRE} for Diffusion-Generated Image Detection}, 
	year={2023},
	pages={22388-22398},
	//keywords={Visualization;Perturbation methods;Training data;Detectors;Image representation;Benchmark testing;Solids},
	doi={10.1109/ICCV51070.2023.02051}}

@inproceedings{zhu2023genimage,
	author = {Zhu, Mingjian and Chen, Hanting and YAN, Qiangyu and Huang, Xudong and Lin, Guanyu and Li, Wei and Tu, Zhijun and Hu, Hailin and Hu, Jie and Wang, Yunhe},
	booktitle = {Advances in Neural Information Processing Systems},
	//editor = {A. Oh and T. Naumann and A. Globerson and K. Saenko and M. Hardt and S. Levine},
	pages = {77771--77782},
	publisher = {Curran Associates, Inc.},
	title = {{GenImage}: A Million-Scale Benchmark for Detecting AI-Generated Image},
	url = {https://proceedings.neurips.cc/paper_files/paper/2023/file/f4d4a021f9051a6c18183b059117e8b5-Paper-Datasets_and_Benchmarks.pdf},
	volume = {36},
	year = {2023}
}

@inproceedings{deng2009imagenet,
	author={Deng, Jia and Dong, Wei and Socher, Richard and Li, Li-Jia and Kai Li and Li Fei-Fei},
	booktitle={IEEE Conference on Computer Vision and Pattern Recognition}, 
	title={{ImageNet}: A Large-Scale Hierarchical Image Database}, 
	year={2009},
	pages={248-255},
	doi={10.1109/CVPR.2009.5206848}
}

@inproceedings{karras2018progressive,
	title={Progressive Growing of {GAN}s for Improved Quality, Stability, and Variation},
	author={Tero Karras and Timo Aila and Samuli Laine and Jaakko Lehtinen},
	booktitle={International Conference on Learning Representations},
	year={2018},
	url={https://openreview.net/forum?id=Hk99zCeAb},
}

@inproceedings{ho2020denoising,
	author = {Ho, Jonathan and Jain, Ajay and Abbeel, Pieter},
	booktitle = {Advances in Neural Information Processing Systems},
	//editor = {H. Larochelle and M. Ranzato and R. Hadsell and M.F. Balcan and H. Lin},
	pages = {6840--6851},
	publisher = {Curran Associates, Inc.},
	title = {Denoising Diffusion Probabilistic Models},
	url ={https://proceedings.neurips.cc/paper_files/paper/2020/file/4c5bcfec8584af0d967f1ab10179ca4b-Paper.pdf},
	volume = {33},
	year = {2020}
}

@preprint{zhong2024patchcraft,
	author         = {Nan Zhong and Yiran Xu and Sheng Li and Zhenxing Qian and Xinpeng Zhang},
	title          = {{PatchCraft}: Exploring Texture Patch for Efficient {AI}-generated Image Detection},
	year           = {2024},
	archivePrefix  = {arXiv},
	eprint         = {2311.12397},
	doi            = {10.48550/arXiv.2311.12397}
}

@inproceedings{liu2022detecting,
	author = {Liu, Bo and Yang, Fan and Bi, Xiuli and Xiao, Bin and Li, Weisheng and Gao, Xinbo},
	title = {Detecting Generated Images by Real Images},
	year = {2022},
	isbn = {978-3-031-19780-2},
	publisher = {Springer-Verlag},
	address = {Berlin, Heidelberg},
	//url = {https://doi.org/10.1007/978-3-031-19781-9_6},
	doi = {10.1007/978-3-031-19781-9_6},
	booktitle = {Proceedings of the European Conference on Computer Vision},
	pages = {95–110},
	//numpages = {16},
	//location = {Tel Aviv, Israel}
}

@article{karras2021style,
	author={Karras, Tero and Laine, Samuli and Aila, Timo},
	journal={IEEE Transactions on Pattern Analysis \& Machine Intelligence },
	title={A Style-Based Generator Architecture for Generative Adversarial Networks},
	year={2021},
	volume={43},
	number={12},
	ISSN={1939-3539},
	pages={4217-4228},
	doi={10.1109/TPAMI.2020.2970919},
	url = {https://doi.ieeecomputersociety.org/10.1109/TPAMI.2020.2970919},
	publisher={IEEE Computer Society},
	//address={Los Alamitos, CA, USA},
	//month=dec
}

@InProceedings{wang2020cnn,
	author={Wang, Sheng-Yu and Wang, Oliver and Zhang, Richard and Owens, Andrew and Efros, Alexei A.},
	booktitle={Proceedings of the IEEE/CVF Conference on Computer Vision and Pattern Recognition}, 
	title={{CNN}-Generated Images Are Surprisingly Easy to Spot… for Now}, 
	year={2020},
	pages={8692-8701},
	//keywords={Gallium nitride;Training;Image generation;Face;Image resolution;Detectors;Generators},
	doi={10.1109/CVPR42600.2020.00872}}

@inproceedings{chollet2017xception,
	author={Chollet, François},
	booktitle={Proceedings of the IEEE Conference on Computer Vision and Pattern Recognition}, 
	title={{Xception}: Deep Learning with Depthwise Separable Convolutions}, 
	year={2017},
	pages={1800-1807},
	//keywords={Computer architecture;Correlation;Convolutional codes;Google;Biological neural networks},
	doi={10.1109/CVPR.2017.195}}

@inproceedings{li2022efficientformer,
	author = {Li, Yanyu and Yuan, Geng and Wen, Yang and Hu, Ju and Evangelidis, Georgios and Tulyakov, Sergey and Wang, Yanzhi and Ren, Jian},
	booktitle = {Advances in Neural Information Processing Systems},
	//editor = {S. Koyejo and S. Mohamed and A. Agarwal and D. Belgrave and K. Cho and A. Oh},
	pages = {12934--12949},
	publisher = {Curran Associates, Inc.},
	title = {{EfficientFormer}: Vision Transformers at MobileNet Speed},
	url = {https://proceedings.neurips.cc/paper_files/paper/2022/file/5452ad8ee6ea6e7dc41db1cbd31ba0b8-Paper-Conference.pdf},
	volume = {35},
	year = {2022}
}

@article{wu2013local,
	title = {Local Shannon Entropy Measure With Statistical Tests for Image Randomness},
	journal = {Information Sciences},
	volume = {222},
	pages = {323-342},
	year = {2013},
	//note = {Including Special Section on New Trends in Ambient Intelligence and Bio-inspired Systems},
	issn = {0020-0255},
	doi = {https://doi.org/10.1016/j.ins.2012.07.049},
	url = {https://www.sciencedirect.com/science/article/pii/S002002551200521X},
	author = {Yue Wu and Yicong Zhou and George Saveriades and Sos Agaian and Joseph P. Noonan and Premkumar Natarajan},
	//keywords = {Image encryption, Shannon entropy, Image randomness, Hypothesis test}
}

@inproceedings{xu2019training,
	author = {Xu, Zhi-Qin John and Zhang, Yaoyu and Xiao, Yanyang},
	title = {Training Behavior of Deep Neural Network in Frequency Domain},
	year = {2019},
	isbn = {978-3-030-36707-7},
	publisher = {Springer-Verlag},
	address = {Berlin, Heidelberg},
	url = {https://doi.org/10.1007/978-3-030-36708-4_22},
	doi = {10.1007/978-3-030-36708-4_22},
	booktitle = {Proceedings of the 26th International Conference on Neural Information Processing},
	pages = {264--274},
	//numpages = {11},
	//keywords = {Generalization, Fourier analysis, Deep learning, Deep Neural Network},
	//location = {Sydney, NSW, Australia}
}

@InProceedings{rahaman2019spectral,
  title = 	 {On the Spectral Bias of Neural Networks},
  author =       {Rahaman, Nasim and Baratin, Aristide and Arpit, Devansh and Draxler, Felix and Lin, Min and Hamprecht, Fred and Bengio, Yoshua and Courville, Aaron},
  booktitle = 	 {Proceedings of the International Conference on Machine Learning},
  pages = 	 {5301--5310},
  year = 	 {2019},
  //editor = 	 {Chaudhuri, Kamalika and Salakhutdinov, Ruslan},
  volume = 	 {97},
  series = 	 {Proceedings of Machine Learning Research},
  //month = 	 {09--15 Jun},
  publisher =    {PMLR},
  //pdf = 	 {http://proceedings.mlr.press/v97/rahaman19a/rahaman19a.pdf},
  url = 	 {https://proceedings.mlr.press/v97/rahaman19a.html},
}

@article{buades2011NLM,
	title   = {{Non-Local Means Denoising}},
	author  = {Buades, Antoni and Coll, Bartomeu and Morel, Jean-Michel},
	journal = {Image Processing On Line},
	volume  = {1},
	pages   = {208--212},
	year    = {2011},
	doi     = {10.5201/ipol.2011.bcm_nlm},
	url    = {https://doi.org/10.5201/ipol.2011.bcm_nlm}
}

@online{flux2024,
	author       = {{Black Forest Labs}},
	title        = {{FLUX}},
	year         = {2024},
	url          = {https://github.com/black-forest-labs/flux},
	month        = aug,
	lastaccessed = {August 1, 2025}
}

@preprint{labs2025flux1kontextflowmatching,
	author         = {{Black Forest Labs}},
	title          = {{FLUX.1 Kontext}: Flow Matching for In-Context Image Generation and Editing in Latent Space},
	year           = {2025},
	archivePrefix  = {arXiv},
	eprint         = {2506.15742},
	doi            = {10.48550/arXiv.2506.15742},
	url            = {https://arxiv.org/abs/2506.15742}
}

@inproceedings{cheng2024diffusion,
	author = {Cheng, Harry and Guo, Yangyang and Wang, Tianyi and Nie, Liqiang and Kankanhalli, Mohan},
	title = {Diffusion Facial Forgery Detection},
	year = {2024},
	isbn = {9798400706868},
	publisher = {Association for Computing Machinery},
	address = {New York, NY, USA},
	//url = {https://doi.org/10.1145/3664647.3680797},
	doi = {10.1145/3664647.3680797},
	booktitle = {Proceedings of the 32nd ACM International Conference on Multimedia},
	pages = {5939–5948},
	//numpages = {10},
	//location = {Melbourne VIC, Australia},
	series = {MM '24}
}

@inproceedings{yan2025sanity,
	title={A Sanity Check for {AI}-generated Image Detection},
	author={Shilin Yan and Ouxiang Li and Jiayin Cai and Yanbin Hao and Xiaolong Jiang and Yao Hu and Weidi Xie},
	booktitle={International Conference on Learning Representations},
	year={2025},
	url={https://openreview.net/forum?id=ODRHZrkOQM}
}

@InProceedings{nichol2021improved,
	title = 	 {Improved Denoising Diffusion Probabilistic Models},
	author =     {Nichol, Alexander Quinn and Dhariwal, Prafulla},
	booktitle =  {Proceedings of the International Conference on Machine Learning},
	pages = 	 {8162--8171},
	year = 	 {2021},
	//editor = 	 {Meila, Marina and Zhang, Tong},
	volume = 	 {139},
	series = 	 {Proceedings of Machine Learning Research},
	//month = 	 {18--24 Jul},
	publisher =    {PMLR},
	//pdf = 	 {http://proceedings.mlr.press/v139/nichol21a/nichol21a.pdf},
	url = 	 {https://proceedings.mlr.press/v139/nichol21a.html},
}

@inproceedings{song2021denoising,
	title={Denoising Diffusion Implicit Models},
	author={Jiaming Song and Chenlin Meng and Stefano Ermon},
	booktitle={International Conference on Learning Representations},
	year={2021},
	url={https://openreview.net/forum?id=St1giarCHLP}
}

@inproceedings{liu2022pseudo,
	title={Pseudo Numerical Methods for Diffusion Models on Manifolds},
	author={Luping Liu and Yi Ren and Zhijie Lin and Zhou Zhao},
	booktitle={International Conference on Learning Representations},
	year={2022},
	url={https://openreview.net/forum?id=PlKWVd2yBkY}
}

@preprint{wang2022diffusiondb,
	author         = {Zijie J. Wang and Evan Montoya and David Munechika and Haoyang Yang and Benjamin Hoover and Duen Horng Chau},
	title          = {{DiffusionDB}: A Large-Scale Prompt Gallery Dataset for Text-to-Image Generative Models},
	year           = {2022},
	archivePrefix  = {arXiv},
	eprint         = {2210.14896},
	doi            = {10.48550/arXiv.2210.14896},
	url            = {https://arxiv.org/abs/2210.14896}
}

@article{zhang2014bp4d,
	title = {{BP4D-Spontaneous}: A High-Resolution Spontaneous 3D Dynamic Facial Expression Database},
	journal = {Image and Vision Computing},
	volume = {32},
	number = {10},
	pages = {692-706},
	year = {2014},
	//note = {Best of Automatic Face and Gesture Recognition 2013},
	issn = {0262-8856},
	doi = {https://doi.org/10.1016/j.imavis.2014.06.002},
	url = {https://www.sciencedirect.com/science/article/pii/S0262885614001012},
	author = {Xing Zhang and Lijun Yin and Jeffrey F. Cohn and Shaun Canavan and Michael Reale and Andy Horowitz and Peng Liu and Jeffrey M. Girard},
	//keywords = {3D facial expression, FACS, Spontaneous expression, Dynamic facial expression database},
}

@inproceedings{corvi2023detection,
	author={Corvi, Riccardo and Cozzolino, Davide and Zingarini, Giada and Poggi, Giovanni and Nagano, Koki and Verdoliva, Luisa},
	booktitle={IEEE International Conference on Acoustics, Speech and Signal Processing}, 
	title={On The Detection of Synthetic Images Generated by Diffusion Models}, 
	year={2023},
	pages={1-5},
	doi={10.1109/ICASSP49357.2023.10095167}
}

@preprint{mccloskey2018detecting_color,
	author         = {Scott McCloskey and Michael Albright},
	title          = {Detecting {GAN}-Generated Imagery Using Color Cues},
	year           = {2018},
	archivePrefix  = {arXiv},
	eprint         = {1812.08247},
	doi            = {10.48550/arXiv.1812.08247},
	url            = {https://arxiv.org/abs/1812.08247}
}

@inproceedings{mccloskey2019detecting_saturation,
	author={McCloskey, Scott and Albright, Michael},
	booktitle={IEEE International Conference on Image Processing}, 
	title={Detecting GAN-Generated Imagery using Saturation Cues}, 
	year={2019},
	pages={4584--4588},
	publisher = {IEEE},
	doi={10.1109/ICIP.2019.8803661}
}

@preprint{nataraj2019detectinggangeneratedfake,
	author         = {Lakshmanan Nataraj and Tajuddin Manhar Mohammed and Shivkumar Chandrasekaran and Arjuna Flenner and Jawadul H. Bappy and Amit K. Roy-Chowdhury and B. S. Manjunath},
	title          = {Detecting {GAN}-Generated Fake Images Using Co-Occurrence Matrices},
	year           = {2019},
	archivePrefix  = {arXiv},
	eprint         = {1903.06836},
	doi            = {10.48550/arXiv.1903.06836},
	url            = {https://arxiv.org/abs/1903.06836}
}

@InProceedings{frank2020leveraging,
	title = 	 {Leveraging Frequency Analysis for Deep Fake Image Recognition},
	author =    {Frank, Joel and Eisenhofer, Thorsten and Sch{\"o}nherr, Lea and Fischer, Asja and Kolossa, Dorothea and Holz, Thorsten},
	booktitle =   {Proceedings of the International Conference on Machine Learning},
	pages = 	 {3247--3258},
	year = 	 {2020},
	//editor = 	 {III, Hal Daumé and Singh, Aarti},
	volume = 	 {119},
	series = 	 {Proceedings of Machine Learning Research},
	//month = 	 {13--18 Jul},
	publisher =    {PMLR},
	//pdf = 	 {http://proceedings.mlr.press/v119/frank20a/frank20a.pdf},
	url = 	 {https://proceedings.mlr.press/v119/frank20a.html},	
}

@inproceedings{ojha2023towards,
	author={Ojha, Utkarsh and Li, Yuheng and Lee, Yong Jae},
	booktitle={Proceedings of the IEEE/CVF Conference on Computer Vision and Pattern Recognition}, 
	title={Towards Universal Fake Image Detectors that Generalize Across Generative Models}, 
	year={2023},
	pages={24480-24489},
	//keywords={Training;Computer vision;Codes;Computational modeling;Buildings;Detectors;Generative adversarial networks;Image and video synthesis and generation},
	doi={10.1109/CVPR52729.2023.02345}}

@preprint{deepseek2024v2,
	author         = {DeepSeek-AI},
	title          = {{DeepSeek-V2}: A Strong, Economical, and Efficient Mixture-of-Experts Language Model},
	year           = {2024},
	archivePrefix  = {arXiv},
	eprint         = {2405.04434},
	doi            = {10.48550/arXiv.2405.04434},
	url            = {https://arxiv.org/abs/2405.04434}
}

@preprint{openai2024gpt4osystemcard,
	author         = {{OpenAI}},
	title          = {GPT-4o System Card},
	year           = {2024},
	archivePrefix  = {arXiv},
	eprint         = {2410.21276},
	doi            = {10.48550/arXiv.2410.21276},
	url            = {https://arxiv.org/abs/2410.21276}
}

@preprint{qwen,
	author         = {Jinze Bai and Shuai Bai and Yunfei Chu and Zeyu Cui and Kai Dang and Xiaodong Deng and Yang Fan and Wenbin Ge and Yu Han and Fei Huang and Binyuan Hui and Luo Ji and Mei Li and Junyang Lin and Runji Lin and Dayiheng Liu and Gao Liu and Chengqiang Lu and Keming Lu and Jianxin Ma and Rui Men and Xingzhang Ren and Xuancheng Ren and Chuanqi Tan and Sinan Tan and Jianhong Tu and Peng Wang and Shijie Wang and Wei Wang and Shengguang Wu and Benfeng Xu and Jin Xu and An Yang and Hao Yang and Jian Yang and Shusheng Yang and Yang Yao and Bowen Yu and Hongyi Yuan and Zheng Yuan and Jianwei Zhang and Xingxuan Zhang and Yichang Zhang and Zhenru Zhang and Chang Zhou and Jingren Zhou and Xiaohuan Zhou and Tianhang Zhu},
	title          = {Qwen Technical Report},
	year           = {2023},
	archivePrefix  = {arXiv},
	eprint         = {2309.16609},
	doi            = {10.48550/arXiv.2309.16609},
	url            = {https://arxiv.org/abs/2309.16609}
}

@online{un_wpp2024,
	author       = {{United Nations, Department of Economic and Social Affairs, Population Division}},
	title        = {World Population Prospects 2024},
	year         = {2024},
	url          = {https://population.un.org/wpp/},
	month        = dec,
	lastaccessed = {December 17, 2024}
}

@inproceedings{han2024face,
	author = {Han, Yue and Zhu, Junwei and He, Keke and Chen, Xu and Ge, Yanhao and Li, Wei and Li, Xiangtai and Zhang, Jiangning and Wang, Chengjie and Liu, Yong},
	title = {Face-Adapter for Pre-trained Diffusion Models with Fine-Grained ID and Attribute Control},
	year = {2024},
	isbn = {978-3-031-72972-0},
	publisher = {Springer-Verlag},
	address = {Berlin, Heidelberg},
	//url = {https://doi.org/10.1007/978-3-031-72973-7_2},
	doi = {10.1007/978-3-031-72973-7_2},
	booktitle = {Proceedings of the European Conference on Computer Vision},
	pages = {20–36},
	//numpages = {17},
	//keywords = {Face Reenactment, Face Swapping, Diffusion Model},
	//location = {Milan, Italy}
}

@InProceedings{yi2023SAAN,
	author={Yi, Ran and Tian, Haoyuan and Gu, Zhihao and Lai, Yu-Kun and Rosin, Paul L.},
	booktitle={Proceedings of the IEEE/CVF Conference on Computer Vision and Pattern Recognition}, 
	title={Towards Artistic Image Aesthetics Assessment: a Large-scale Dataset and a New Method}, 
	year={2023},
	pages={22388-22397},
	//keywords={Photography;Computer vision;Art;Codes;Current measurement;Computational modeling;Benchmark testing;Datasets and evaluation},
	doi={10.1109/CVPR52729.2023.02144}}

@InProceedings{he2016deep,
	author={He, Kaiming and Zhang, Xiangyu and Ren, Shaoqing and Sun, Jian},
	booktitle={IEEE Conference on Computer Vision and Pattern Recognition}, 
	title={Deep Residual Learning for Image Recognition}, 
	year={2016},
	pages={770-778},
	doi={10.1109/CVPR.2016.90}}

@InProceedings{tan2024NPR,
	author={Tan, Chuangchuang and Liu, Huan and Zhao, Yao and Wei, Shikui and Gu, Guanghua and Liu, Ping and Wei, Yunchao},
	booktitle={Proceedings of the IEEE/CVF Conference on Computer Vision and Pattern Recognition}, 
	title={Rethinking the Up-Sampling Operations in CNN-Based Generative Network for Generalizable Deepfake Detection}, 
	year={2024},
	pages={28130--28139},
	//keywords={Deepfakes;Computer vision;Frequency-domain analysis;Computer architecture;Detectors;Generative adversarial networks;Generators;Deepfake Detection;Up-Sampling Operations;Neighboring Pixel Relationships},
	doi={10.1109/CVPR52733.2024.02657}}

@InProceedings{liu2024fatformer,
	author={Liu, Huan and Tan, Zichang and Tan, Chuangchuang and Wei, Yunchao and Wang, Jingdong and Zhao, Yao},
	booktitle={Proceedings of the IEEE/CVF Conference on Computer Vision and Pattern Recognition}, 
	title={Forgery-aware Adaptive Transformer for Generalizable Synthetic Image Detection}, 
	year={2024},
	pages={10770-10780},
	//keywords={Training;Couplings;Adaptation models;Detectors;Transformer cores;Transformers;Feature extraction;Synthetic Image Detection;Forgery Adaptation},
	doi={10.1109/CVPR52733.2024.01024}}

\end{document}